\documentclass[10pt,twocolumn,letterpaper]{article}
\usepackage[pagenumbers]{cvpr}

\usepackage{multirow}
\usepackage{multicol}

\usepackage[dvipsnames]{xcolor}

\definecolor{cvprblue}{rgb}{0.21,0.49,0.74}
\usepackage[pagebackref,breaklinks,colorlinks,citecolor=cvprblue]{hyperref}
\usepackage{xcolor}
\usepackage{colortbl}
\usepackage{graphicx}
\usepackage{tikz} 
\definecolor{deepgreen}{rgb}{0,0.5,0}

\def\paperID{*****} 
\def\confName{CVPR}
\def\confYear{2024}

\newcommand{\ourmodel}{\textit{T-Rex2}}

\title{T-Rex2: Towards Generic Object Detection via Text-Visual Prompt Synergy}

\author{Qing Jiang$^{1,2\S}$ , Feng Li$^{2,3\S}$ , Zhaoyang Zeng$^{2}$ , Tianhe Ren$^{2}$ , Shilong Liu$^{2,4\S}$ , Lei Zhang$^{1,2\dagger}$ \\
$^1$South China University of Technology \\ 
$^2$International Digital Economy Academy (IDEA) \\
$^3$The Hong Kong University of Science and Technology \\ $^4$Tsinghua University\\
{\tt\small mountchicken@outlook.com , fliay@connect.ust.hk , lius120@mails.tsinghua.edu.cn} \\ {\tt\small \{rentianhe, zengzhaoyang, leizhang\}@idea.edu.cn} \\
\url{https://deepdataspace.com/home}
}

\begin{document}
\twocolumn[{
\maketitle\centering
\captionsetup{type=figure}
\includegraphics[width=0.9\textwidth]{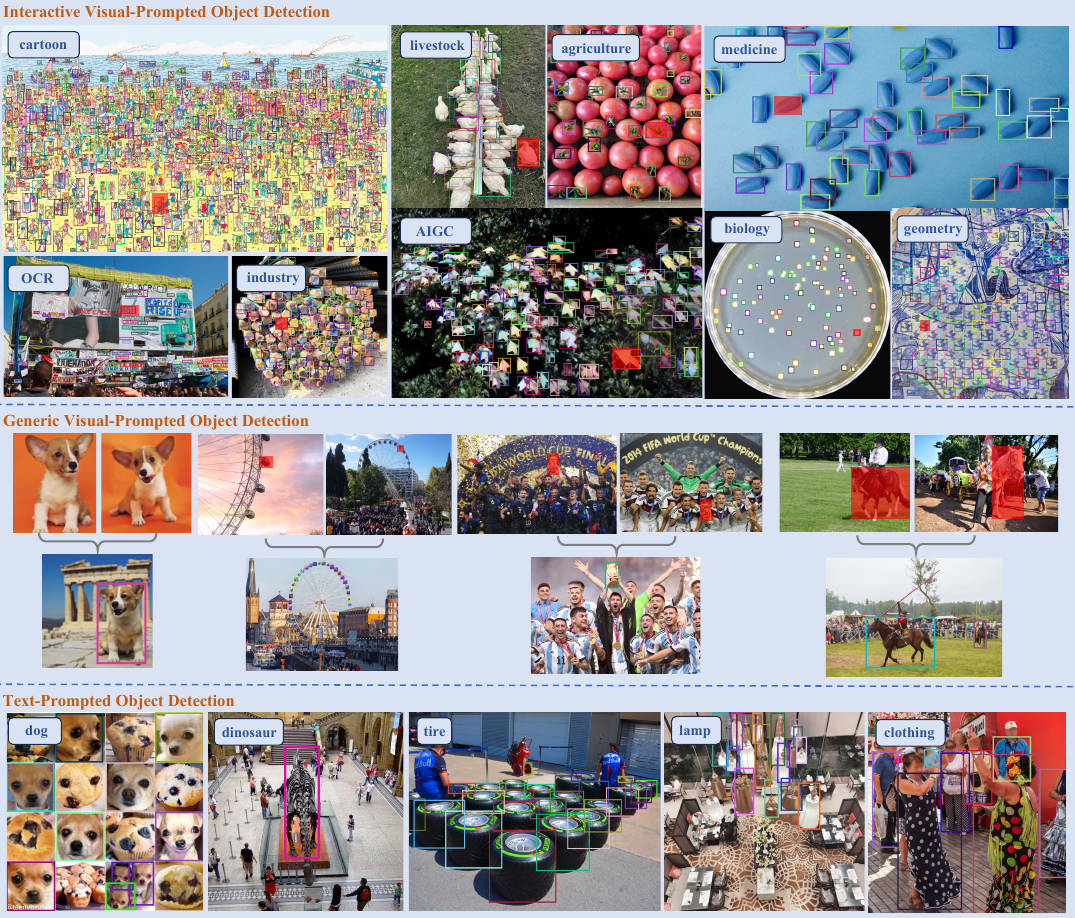}
\captionof{figure}{We introduce \ourmodel, a promptable and interactive model for open-set object detection. \ourmodel{} can take both text prompts and visual prompts (boxes or points within the same image or across multiple images) as input for object detection. \ourmodel{} delivers strong zero-shot object detection capabilities and is highly practical for various scenarios, with \textbf{only one suit of weights.}}
\label{fig:teaser}
}]
\makeatletter
\def\@makefnmark{} 
\makeatother
\footnotetext{\S\emph{This work was done when Qing Jiang, Feng Li, and Shilong Liu were interns at IDEA.}}

\footnotemark
\addtocounter{footnote}{-1}
\footnotetext{\emph{$\dagger$ Corresponding author.}}

\maketitle
\begin{abstract}
We present \ourmodel{}, a highly practical model for open-set object detection. Previous open-set object detection methods relying on text prompts effectively encapsulate the abstract concept of common objects, but struggle with rare or complex object representation due to data scarcity and descriptive limitations. Conversely, visual prompts excel in depicting novel objects through concrete visual examples, but fall short in conveying the abstract concept of objects as effectively as text prompts. Recognizing the complementary strengths and weaknesses of both text and visual prompts, we introduce \ourmodel{} that synergizes both prompts within a single model through contrastive learning. \ourmodel{} accepts inputs in diverse formats, including text prompts, visual prompts, and the combination of both, so that it can handle different scenarios by switching between the two prompt modalities. Comprehensive experiments demonstrate that \ourmodel{} exhibits remarkable zero-shot object detection capabilities across a wide spectrum of scenarios. We show that text prompts and visual prompts can benefit from each other within the synergy, which is essential to cover massive and complicated real-world scenarios and pave the way towards generic object detection. Model API is now available at \url{https://github.com/IDEA-Research/T-Rex}.

\end{abstract}    
\section{Introduction}
\label{sec:intro}

Object detection, a foundational pillar of computer vision, aims to locate and identify objects within an image. Traditionally, object detection was operated within a closed-set paradigm~\cite{girshick2015fast, ren2015faster, redmon2016you, lin2017focal, tian2019fcos,zhou2019objects, zhu2020deformable, liu2022dabdetr, li2022dn, zhang2022dino, carion2020end}, wherein a predefined set of categories is known a prior, and the system is trained to recognize and detect objects from this set. Yet the ever-changing and unforeseeable nature of the real world demands a shift in object detection methodologies towards an open-set paradigm.

Open-set object detection represents a significant paradigm shift, transcending the limitations of closed-set detection by empowering models to identify objects beyond a predetermined set of categories. A prevalent approach is to use text prompts for open-vocabulary object detection~\cite{liu2023grounding, li2022grounded, ghiasi2022scaling, zhou2022detecting, gu2021open, minderer2022simple, kamath2021mdetr}. This approach typically involves distilling knowledge from language models like CLIP~\cite{clip} or BERT~\cite{devlin2018bert} to align textual descriptions with visual representations.                

\begin{figure}[t]
\centering
\includegraphics[width=0.95\linewidth]{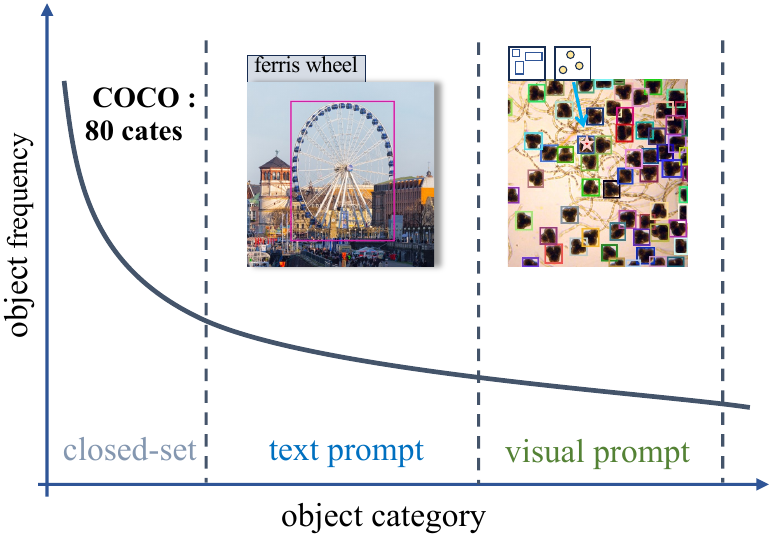}\vspace{-1mm}
\caption{Long-tailed curve of object frequency and the number of categories that can be detected. We suggest that the text prompt can cover the middle part of the long-tailed curve, while the visual prompt can cover the tail.}
\label{fig:intro1} 
\vspace{-2mm}
\end{figure}

While using text prompts has been predominantly favored in open-set detection for their capacity to abstractly describe objects, it still faces the following limitations. 
1) \textit{Long-tailed data shortage}. The training of text prompts necessitates modality alignment between visual representations, however, the scarcity of data for long-tailed objects may impair the learning efficiency. As depicted in ~\cref{fig:intro1}, the distribution of objects inherently follows a long-tail pattern, i.e., as the variety of detectable objects increases, the available data for these objects becomes increasingly scarce. This data scarcity may undermine the capacity of models to identify rare or novel objects. 2) \textit{Descriptive limitations}. Text prompts also fall short of accurately depicting objects that are hard to describe in language. For instance, as shown in~\cref{fig:intro1}, while a text prompt may effectively describe \texttt{ferris wheel}, it may struggle to accurately represent \texttt{the microorganisms in the microscope image} without biological knowledge.

 Conversely, visual prompts~\cite{xu2023multi, zou2023segment, li2023semantic, kirillov2023segment, li2023visual, jiang2023t} provide a more intuitive and direct method to represent objects by providing visual examples. For example, users can use points or boxes to mark the object for detection, even if they do not know what the object is. Additionally, visual prompts are not constrained by the need for cross-modal alignment, since they rely on visual similarity rather than linguistic correlation, enabling their application to novel objects that are not encountered during training.  

 Nonetheless, visual prompts also exhibit limitations, as they are less effective at capturing the general concept of objects compared to text prompts. For instance, the term \texttt{dog} as a text prompt broadly covers all dog varieties. In contrast, visual prompts, given the vast diversity in dog breeds, sizes, and colors, would necessitate a comprehensive image collection to visually convey the abstract notion of \texttt{dog}.

Recognizing the complementary strengths and weaknesses of both text and visual prompts, we introduce \ourmodel{}, a generic open-set object detection model that integrates both modalities. \ourmodel{} is built upon the DETR~\cite{carion2020end} architecture which is an end-to-end object detection model. It incorporates two parallel encoders to encode both text and visual prompts. For text prompts, we utilize the text encoder of CLIP~\cite{clip} to encode input text into text embedding. For visual prompts, we introduce a novel visual prompt encoder equipped with the deformable attention~\cite{zhu2020deformable} that can transform the input visual prompts (points or boxes) on a single image or across multiple images into visual prompt embeddings. To facilitate the collaboration of these two prompt modalities, we propose a contrastive learning~\cite{clip, hadsell2006dimensionality} module that can explicitly align text prompts and visual prompts. During the alignment, visual prompts can benefit from the generality and abstraction capabilities inherent in text prompts. Conversely, text prompts can enhance their descriptive capabilities by looking at various visual prompts. This iterative interaction allows both visual and text prompts to evolve continuously, thereby improving their ability for generic understanding within one model.

\ourmodel{} supports four unique workflows that can be applied to various scenarios: 1) \textit{interactive visual prompt workflow}, allowing users to specify the object to be detected by given visual example through box or point on the current image; 2) \textit{generic visual prompt workflow}, permitting users to define a specific object across multiple images through visual prompts, thereby creating a universal visual embedding applicable to other images; 3) \textit{text prompt workflow}, enabling users to employ descriptive text for open-vocabulary object detection; 4) \textit{mix prompt workflow}, which combines both text and visual prompts for joint inference.

\ourmodel{} demonstrates strong object detection capabilities and achieves remarkable results on COCO~\cite{lin2014microsoft}, LVIS~\cite{gupta2019lvis}, ODinW~\cite{li2022elevater} and Roboflow100~\cite{ciaglia2022roboflow}, all under zero-shot setting. Through our analysis, we observe that text and visual prompts serve complementary roles, each excelling in scenarios where the other may not be as effective. Specifically, text prompts are particularly good at recognizing common objects, while visual prompts excel in rare objects or scenarios that may not be easily described through language. This complementary relationship enables the model to perform effectively across a wide range of scenarios. To summarize, our contributions are threefold:
\begin{itemize}
    \item We propose an open-set object detection model \ourmodel{} that unifies text and visual prompts within one framework, which demonstrates strong zero-shot capabilities across various scenarios.
    \item We propose a contrastive learning module to explicitly align text and visual prompts, which leads to mutual enhancement of these two modalities.
    \item Extensive experiments demonstrate the benefits of unifying text and visual prompts within one model. We also reveal that each type of prompt can cover different scenarios, which collectively show promise in advancing toward general object detection.
\end{itemize}
\section{Related Work}
\label{sec:related_work}
\subsection{Text-prompted Object Detection}
Remarkable progress has been achieved in text-prompted object detection~\cite{liu2023grounding, gu2021open, zang2022open, zhong2022regionclip, zhang2023simple, minderer2022simple, li2022grounded, kamath2021mdetr}, which demonstrate impressive zero-shot and few-shot recognition capabilities. These models are typically built upon a pre-trained text encoder like CLIP~\cite{clip} and BERT~\cite{devlin2018bert}. GLIP~\cite{li2022grounded} proposes to formulate object detection as grounding problems, which unifies different data formats to align different modalities and expand detection vocabulary. Following GLIP, Grounding DINO~\cite{liu2023grounding} improves the vision-language alignment by fusing different modalities in the early phase. DetCLIP~\cite{yao2022detclip} and RegionCLIP~\cite{zhong2022regionclip} leverages image-text pairs with pseudo boxes to expand region knowledge for more generalized object detection.

\subsection{Visual-prompted Object Detection}
Beyond text-prompted models, developing models incorporating visual prompts is a trending research area due to its flexibility and context-awareness. Mainstream visual-prompted models~\cite{zang2022open, xu2023multi, minderer2022simple} adopt raw images as visual prompts and leverage image-text-aligned representation to transfer knowledge from text to visual prompts. However, it is restricted to image-level prompts and highly relies on aligned image-text foundation models. Another emergent approach for visual prompts is to use visual instructions like box, point, and referred region of another image. DINOv~\cite{li2023visual} proposes to use visual prompts as in-context examples for open-set detection and segmentation tasks. When detecting a novel category, it takes in several visual examples of this category to understand this category in an in-context manner. In this paper, we focus on visual prompts in the form of visual instructions. 

\begin{figure*}[h]\centering
\includegraphics[width=0.8\linewidth]{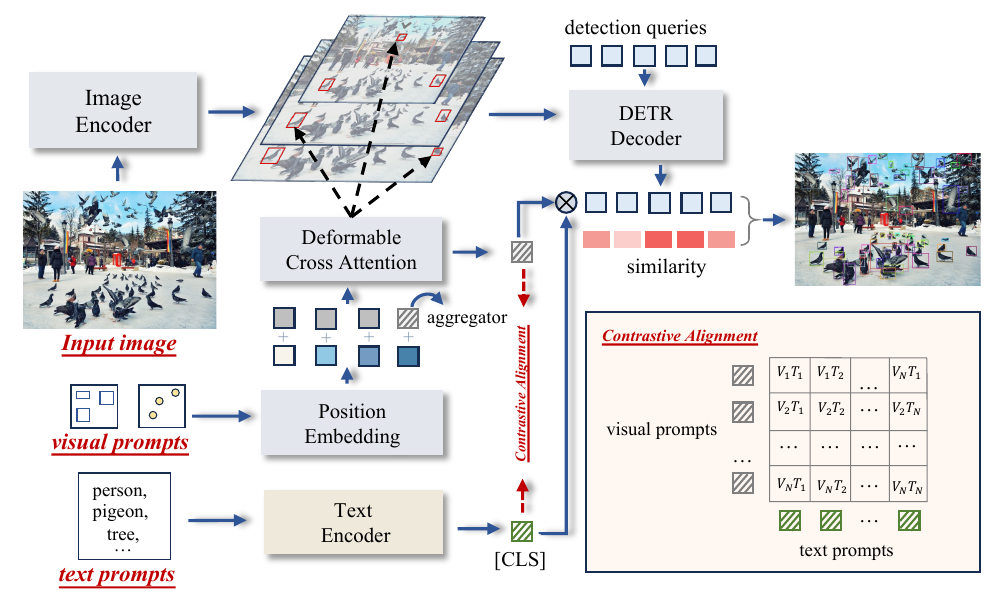}\vspace{-1mm}
\caption{Overview of the \ourmodel{} model. \ourmodel{} mainly follows the design principles of DETR~\cite{carion2020end} which is an end-to-end object detection model. Visual prompt and text prompt are introduced through deformable cross attention~\cite{zhu2020deformable} and CLIP~\cite{clip} text encoder, respectively, and are aligned through contrastive learning. }
\label{fig:T-Rex++Model}
\vspace{-1mm}
\end{figure*}

\subsection{Interactive Object Detection} 
Interactive models have shown significant promise in aligning human intentions in the field of computer vision. It has been wildly applied for interactive segmentation~\cite{kirillov2023segment, li2023semantic, zou2023segment}, where the user provides a visual prompt (box, point, and mask, \etc.) and the model outputs a mask corresponding to the prompt. This process typically follows a one-to-one interaction model, i.e., one prompt for one output mask. However, object detection requires a one-to-many approach, where a single visual prompt can lead to multiple detected boxes. Several works~\cite{yao2012interactive,lee2022interactive} have incorporated interactive object detection for the purpose of automating annotations. T-Rex~\cite{jiang2023t} leverages interactive visual prompts for the task of object counting through object detection, however, its capabilities in generic object detection have not been extensively explored.

\section{\ourmodel{} Model}
\ourmodel{} integrates four components, as illustrated in Fig. \ref{fig:T-Rex++Model}: i) Image Encoder, ii) Visual Prompt Encoder, iii) Text Prompt Encoder, and iv) Box Decoder. \ourmodel{} adheres to the design principles of DETR~\cite{carion2020end} which is an end-to-end object detection model. These four components collectively facilitate four distinct workflows that encompass a broad range of application scenarios.

\subsection{Visual-Text Promptable Object Detection}

\noindent\textbf{Image Encoder.} Mirroring the Deformable DETR~\cite{zhu2020deformable} framework, the image encoder in \ourmodel{} consists of a vision backbone (e.g. Swin Transformer~\cite{liu2021swin}) that extracts multi-scale feature maps from input image. This is followed by several transformer encoder layers~\cite{dosovitskiy2020image} equipped with deformable self-attention~\cite{zhu2020deformable}, which are utilized to refine these extracted feature maps. The feature maps output from the image encoder is denoted as $\boldsymbol{f_i} \in \mathbb{R}^{C_i \times H_i \times W_i}, i \in \{1,2,...,L\}$, where $L$ is the number of feature map layers.

\noindent\textbf{Visual Prompt Encoder.} Visual prompt has been widely used in interactive segmentation~\cite{li2023semantic, zou2023segment, kirillov2023segment}, yet to be fully explored within the domain of object detection. Our method incorporates visual prompts in both box and point formats. The design principle involves transforming user-specified visual prompts from their coordinate space to the image feature space.  Given $K$ user-specified 4D normalized boxes $b_j=(x_j,y_j,w_j,h_j), j \in \{1,2,...,K\}$, or 2D normalized points $p_j=(x_j,y_j), j \in \{1,2,...,K\}$ on a reference image, we initially encode these coordinate inputs into position embeddings through a fixed sine-cosine embedding layer. Subsequently, two distinct linear layers are employed to project these embeddings into a uniform dimension:
\begin{gather}
    B = \operatorname{Linear}(\operatorname{PE}(b_1,...b_K); \theta_B) : \mathbb{R}^{K \times 4D} \to \mathbb{R}^{K \times D} \\
    P = \operatorname{Linear}(\operatorname{PE}(p_1,...p_K); \theta_P) : \mathbb{R}^{K \times 2D} \to \mathbb{R}^{K \times D}
\end{gather}

where $\operatorname{PE}$ stands for position embedding and $\operatorname{Linear}(\cdot; \theta)$ indicate a linear project operation with parameter $\theta$. Different from the previous method~\cite{li2023semantic} that regards point as a box with minimal width and height, we model box and point as distinct prompt types. We then initiate a learnable content embedding that is broadcasted $K$ times, denoted as $C \in \mathbb{R}^{K \times D}$. Additionally, a universal class token $C' \in \mathbb{R}^{1 \times D}$ is utilized to aggregate features from other visual prompts, accommodating the scenario where users might supply multiple visual prompts within a single image. These content embeddings are concatenated with position embeddings along the channel dimension, and a linear layer is applied for projection, thereby constructing the input query embedding $Q$:
\begin{equation}
Q = \begin{cases} 
\operatorname{Linear}\left(\operatorname{CAT}\left(\left[C; C'\right], \left[B; B'\right]\right); \varphi_B\right) , \operatorname{box} \\
\operatorname{Linear}\left(\operatorname{CAT}\left(\left[C; C'\right], \left[P; P'\right]\right); \varphi_P\right), \operatorname{ point}
\end{cases} 
\end{equation}

where notion $\operatorname{CAT}$ stands for concatenation at channel dimension. $B'$ and $P'$ represent global position embeddings, which are derived from global normalized coordinates $[0.5, 0.5, 1, 1]$ and $[0.5, 0.5]$. The global query serves the purpose of aggregating features from other queries. Subsequently, we employ a multi-scale deformable cross-attention~\cite{zhu2020deformable} layer to extract visual prompt features from the multi-scale feature maps, conditioned on the visual prompts. For the $j$-th prompt, the query feature ${Q'_j}$ after cross attention is computed as:
\begin{equation}
{Q'_j} = \begin{cases} 
\operatorname{MSDeformAttn}(Q_j, b_j, \left\{\boldsymbol{f}_i\right\}_{i=1}^L) , \operatorname{box} \\
\operatorname{MSDeformAttn}(Q_j, p_j, \left\{\boldsymbol{f}_i\right\}_{i=1}^L) , \operatorname{point}
\end{cases} 
\end{equation}

Deformable attention~\cite{zhu2020deformable} was initially employed to address the slow convergence problem encountered in DETR~\cite{carion2020end}. In our approach, we condition deformable attention on the coordinates of visual prompts, i.e., each query will selectively attend to a limited set of multi-scale image features encompassing the regions surrounding the visual prompts. This ensures the capture of visual prompt embeddings representing the objects of interest. Following the extraction process, we use a self-attention layer to regulate the relationships among different queries and a feed-forward layer for projection. The output of the global content query will be used as the final visual prompt embedding $V$.
\begin{equation}
    V=\operatorname{FFN}(\operatorname{SelfAttn}(Q'))[-1]
\end{equation}

\noindent\textbf{Text Prompt Encoder.} We employ the text encoder of CLIP~\cite{clip} to encode category names or short phrases and use the \texttt{[CLS]} token output as the text prompt embedding, denoted as $T$. 

\noindent\textbf{Box Decoder.} We employ a DETR-like decoder for box prediction. Following DINO~\cite{zhang2022dino}, each query is formulated as a 4D anchor coordinate and undergoes iterative refinement across decoder layers. We employ the query selection layer proposed in Grounding DINO~\cite{liu2023grounding} to initialize the anchor coordinates $(x, y, w, h)$. Specifically, We compute the similarity between the encoder feature and the prompt embeddings and select indices with similarity of top 900 to initialize the position embeddings. Subsequently, the detection queries utilize deformable cross-attention~\cite{zhu2020deformable} to focus on the encoded multi-scale image features and are used to predict anchor offsets $(\Delta x, \Delta y, \Delta w, \Delta h)$ at each decoder layer. The final predicted boxes are obtained by summing the anchors and offsets:
\begin{gather}
    (\Delta x, \Delta y, \Delta w, \Delta h) = \operatorname{MLP}(Q_{dec}) \\
    \operatorname{Box} = (x+\Delta x, y+\Delta y, w+\Delta w, h+\Delta h)
\end{gather}
Where $Q_{dec}$ are predicted queries from the box decoder. Instead of using a learnable linear layer to predict class labels, following previous open-set object detection methods~\cite{li2022grounded, liu2023grounding}, we utilize the prompt embeddings as the weights for the classification layer:
\begin{equation}
    \operatorname{Label} = V \cdot Q_{dec}^T : \mathbb{R}^{C \times D} \times \mathbb{R}^{D \times N}  \rightarrow \mathbb{R}^{C \times N}
\end{equation}

Where $C$ denotes the total number of visual prompt classes, and $N$ represents the number of detection queries.

Both visual prompt object detection and open-vocabulary object detection tasks share the same image encoder and box decoder.

\subsection{Region-Level Contrastive Alignment}
To integrate both visual prompt and text prompt within one model, we employ region-level contrastive learning to align these two modalities. Specifically, given an input image and $K$ visual prompt embeddings $V=(v_1, ..., v_K)$ extracted from the visual prompt encoder, along with the text prompt embeddings $T=(t_1, ..., t_K)$ for each prompt region, we calculate the InfoNCE loss~\cite{oord2018representation} between the two types of embeddings:
\begin{equation}
    \mathcal{L}_{align} = -\frac{1}{K} \sum_{i=1}^K \log \frac{\exp(v_i \cdot t_i)}{\sum_{j=1}^K \exp(v_i \cdot t_j)}
\end{equation}

The contrastive alignment can be regarded as a mutual distillation process, whereby each modality contributes to and benefits from the exchange of knowledge. Specifically, text prompts can be seen as a conceptual anchor, around which diverse visual prompts can converge so that the visual prompt can gain general knowledge. Conversely, the visual prompts act as a continuous source of refinement for text prompts. Through exposure to a wide array of visual instances, the text prompt is dynamically updated and enhanced, gaining depth and nuance.

\subsection{Training Strategy and Objective}
\textbf{Visual prompt training strategy.} For visual prompt training, we adopt the strategy of ``current image prompt, current image detect''. Specifically, for each category in a training set image, we randomly choose between one to all available GT boxes to use as visual prompts. We convert these GT boxes into their center point with a 50\% chance for point prompt training. While using visual prompts from different images for cross-image detection training might seem more effective, creating such image pairs poses challenges in an open-set scenario due to inconsistent label spaces across datasets. Despite its simplicity, our straightforward training strategy still leads to strong generalization capability.

\noindent\textbf{Text prompt training strategy.} \ourmodel{} uses both detection data and grounding data for text prompt training. For detection data, we use the category names in the current image as the positive text prompt and randomly sample negative text prompts in the remaining categories. For grounding data, we extract positive phrases corresponding to the bounding boxes and exclude other words in the caption for text input. Following the methodology of DetCLIP~\cite{yao2022detclip, yao2023detclipv2}, we maintain a global dictionary to sample negative text prompts for grounding data, which are concatenated with the positive text prompts. This global dictionary is constructed by selecting the category names and phrase names that occur more than 100 times in the text prompt training data. 

\noindent\textbf{Training objective.} We employ the L1 loss and GIOU~\cite{rezatofighi2019generalized} loss for box regression. For classification loss, following Grounding DINO~\cite{liu2023grounding}, we apply a contrastive loss that measures the difference between the predicted objects and the prompt embeddings. Specifically, we calculate the similarity between each detection query and the visual prompt or text prompt embeddings through a dot product to predict logits, followed by the computation of a sigmoid focal loss~\cite{lin2017focal} for each logit. The box regression and classification loss are initially employed for bipartite matching~\cite{carion2020end} between predictions and ground truths. Subsequently, we calculate the final losses between ground truths and matched predictions, incorporating the same loss components. We use auxiliary loss after each decoder layer and after the encoder outputs. Following DINO~\cite{zhang2022dino}, we also use denoising training to accelerate convergence. The final loss takes the following form:
\begin{equation}
\mathcal{L}_{\text{total}} = \mathcal{L}_{\text{cls}} + \mathcal{L}_{\text{L1}} + \mathcal{L}_{\text{GIoU}} + \mathcal{L}_{\text{DN}} + \mathcal{L}_{\text{align}}
\end{equation}
We adopt a cyclical training strategy that alternates between text prompts and visual prompts in successive iterations.

\subsection{Four Inference Workflows}
\ourmodel{} offers four different workflows by combining text prompts and visual prompts in different ways.

\noindent\textbf{Text prompt workflow.} This workflow exclusively employs text prompts for object detection, which is the same as open-vocabulary object detection. This workflow is suitable for the detection of common objects, where the text prompt can provide clear descriptions.

\noindent\textbf{Interactive visual prompt workflow.} This workflow is designed around a core principle of user-driven interactivity. Given the initial output of \ourmodel{} from the user-provided prompts, users can refine the detection results by adding additional prompts on missed or falsely-detected objects, based on the visualization result. This iterative cycle allows users to fine-tune \ourmodel{}'s performance interactively, ensuring precise detection. This interactive process remains fast and resource-efficient, as \ourmodel{} is a late fusion model that only requires the image encoder to forward once.

\noindent\textbf{Generic Visual Prompt Workflow.} In this workflow, users can customize visual embeddings for specific objects by prompting \ourmodel{} with an arbitrary number of example images. This capability is crucial for generic object detection since a class of object may have very diverse instances, thus we need a certain amount of visual examples to represent it. Let ${V_1}, {V_2},...,{V_n}$, represent the visual embeddings obtained from $n$ different images, the generic visual embeddings $V$ are computed as the mean of these embeddings.
\begin{equation}
    V = \frac{1}{n} \sum_{i=1}^{n} V_i
\end{equation}

\noindent\textbf{Mixed Prompt Workflow.} After the alignment between visual and text prompts, they can be used for inference at the same time. This fusion is achieved by averaging their respective embeddings.
\begin{equation}
    P_{\text{mixed}} = \frac{T + V}{2}
\end{equation}
In this workflow, text prompts contribute to a broad contextual understanding, while visual prompts add precision and concrete visual cues.

\section{Experiments}
\subsection{Data Engines}
For each modality, specialized data engines are designed to curate data suitable for their respective training needs.

\noindent\textbf{Data engine for text prompt.} \ourmodel{} supports the integration of both detection and grounding data for training. Following~\cite{liu2023grounding, li2022grounded}, we utilize detection datasets Objects365~\cite{shao2019objects365}, OpenImages~\cite{krasin2017openimages}, along with the grounding dataset GoldG~\cite{kamath2021mdetr} for training. To enhance the text prompt capabilities of \ourmodel{}, we also make extensive use of pseudo-labeled data from image caption datasets and image classification datasets. Specifically, for image caption data in Conceptual Captions~\cite{sharma2018conceptual} and LAION400M~\cite{schuhmann2021laion} datasets, we use spaCy\footnote{\url{https://spacy.io/}} to extract noun chunks from image captions and use these noun chunks to prompt Grounding DINO~\cite{liu2023grounding} to get boxes. For image classification data in the Bamboo~\cite{zhang2022bamboo} dataset, we simply use the category of the current image to prompt Grounding DINO~\cite{liu2023grounding}. In total, we use 3.15M labeled images and 3.39M pseudo-labeled images for text prompt training.

\begin{table*}[t]
  \resizebox{\linewidth}{!}{
    \begin{tabular}{ccc|c|cccccccc|cc|c}
      \cline{1-15}
      \multirow{3}{*}{Method} & \multirow{3}{*}{\begin{tabular}[c]{@{}c@{}}Prompt \\ Type\end{tabular}} & \multirow{3}{*}{Backbone} & \begin{tabular}[c]{@{}c@{}}COCO-Val\\ Zero-Shot\end{tabular} & \multicolumn{8}{c|}{\begin{tabular}[c]{@{}c@{}}LVIS \\ Zero-Shot\end{tabular}}                                                                     & \multicolumn{2}{c|}{\begin{tabular}[c]{@{}c@{}}ODinW\\ Zero-Shot\end{tabular}} & \begin{tabular}[c]{@{}c@{}}Roboflow100\\ Zero-Shot\end{tabular} \\ \cline{4-15} 
                              &                                                                         &                           & val-80                                                       & \multicolumn{4}{c|}{minival-804}                                                   & \multicolumn{4}{c|}{val-1203}                                 & \multicolumn{2}{c|}{35val}                                                     & 100val                                                          \\ \cline{4-15} 
                              &                                                                         &                           & AP                                                           & AP            & $AP_f$         & $AP_c$         & \multicolumn{1}{c|}{$AP_{r}$}          & AP            & $AP_f$         & $AP_c$         & $AP_r$         & $AP_{avg}$                                & $AP_{med}$                               & $AP_{avg}$                                                         \\ \cline{1-15}
      GLIP-T~\cite{li2022grounded}                  & Text                                                                    & Swin-T                    & 46.7                                                         & 26.0          & 31.0          & 21.4          & \multicolumn{1}{c|}{20.8}          & 17.2          & 25.5          & 12.5          & 10.1          & 19.6                                   & 5.1                                   & -                                                               \\
      GLIP-L~\cite{li2022grounded}                  & Text                                                                    & Swin-L                    & 49.8                                                         & 37.3          & 41.5          & 34.3          & \multicolumn{1}{c|}{28.2}          & 26.9          & 35.4          & 23.3          & 17.1          & 23.4                                   & 11.0                                  & 8.6                                                             \\
      Grounding DINO~\cite{liu2023grounding}          & Text                                                                    & Swin-T                    & 48.4                                                         & 27.4          & 32.7          & 23.3          & \multicolumn{1}{c|}{18.1}          & -             & -             & -             & -             & 22.3                                   & 11.9                                  & -                                                               \\
      Grounding DINO~\cite{liu2023grounding}          & Text                                                                    & Swin-L                    & \textbf{52.5}                                                & 33.9          & 38.8          & 30.7          & \multicolumn{1}{c|}{22.2}          & -             & -             & -             & -             & 26.1                                   & 18.4                                  & -                                                               \\
      DetCLIPv2~\cite{yao2023detclipv2}               & Text                                                                    & Swin-T                    & -                                                            & 40.4          & 40.0          & 41.7          & \multicolumn{1}{c|}{36.0}          & -             & -             & -             & -             & -                                      & -                                     & -                                                               \\
      DetCLIPv2~\cite{yao2023detclipv2}               & Text                                                                    & Swin-L                    & -                                                            & 44.7          & 43.7          & 46.3          & \multicolumn{1}{c|}{43.1}          & -             & -             & -             & -             & -                                      & -                                     & -                                                               \\
      DINOv~\cite{li2023visual}                   & Visual-G                                                                & Swin-T                    & -                                                            & -             & -             & -             & \multicolumn{1}{c|}{-}             & -             & -             & -             & -             & 14.9                                   & 5.4                                   & -                                                               \\
      DINOv~\cite{li2023visual}                   & Visual-G                                                                & Swin-L                    & -                                                            & -             & -             & -             & \multicolumn{1}{c|}{-}             & -             & -             & -             & -             & 15.7                                   & 4.8                                   & -                                                               \\ \hline
      T-Rex2                  & Text                                                                    & Swin-T                    & \cellcolor{red!5}45.8                                        & \cellcolor{red!5}42.8          & \cellcolor{red!5}46.5          & \cellcolor{red!5}39.7          & \multicolumn{1}{c|}{\cellcolor{red!5}37.4}          & \cellcolor{red!5}34.8          & \cellcolor{red!5}41.2          & \cellcolor{red!5}31.5          & \cellcolor{green!5}29.0          & \cellcolor{green!5}18.0                                   & \cellcolor{green!5}4.7                                   & \cellcolor{green!5}8.2                                                             \\
      T-Rex2                  & Visual-G                                                                & Swin-T                    & \cellcolor{red!5}38.8                                        & \cellcolor{red!5}37.4          & \cellcolor{red!5}41.8          & \cellcolor{red!5}33.9          & \multicolumn{1}{c|}{\cellcolor{red!5}29.9}          & \cellcolor{red!5}\underline{34.9}          & \cellcolor{red!5}41.1          & \cellcolor{red!5}30.3          & \cellcolor{green!5}\underline{32.4}          & \cellcolor{green!5}\underline{23.6}                                   & \cellcolor{green!5}\underline{17.5}                                  & \cellcolor{green!5}\underline{17.4}                                                            \\ \hline
      T-Rex2                  & Text                                                                    & Swin-L                    & \cellcolor{red!5}\underline{52.2}                                        & \cellcolor{red!5}\textbf{54.9} & \cellcolor{red!5}\textbf{56.1} & \cellcolor{red!5}\textbf{54.8} & \multicolumn{1}{c|}{\cellcolor{red!5}\textbf{49.2}} & \cellcolor{red!5}\textbf{45.8} & \cellcolor{red!5}\textbf{50.2} & \cellcolor{red!5}\textbf{43.2} & \cellcolor{green!5}42.7          & \cellcolor{green!5}22.0                                   & \cellcolor{green!5}7.3                                   & \cellcolor{green!5}10.5                                                            \\
      T-Rex2                  & Visual-G                                                                & Swin-L                    & \cellcolor{red!5}46.5                                        & \cellcolor{red!5}47.6          & \cellcolor{red!5}49.5          & \cellcolor{red!5}46.0          & \multicolumn{1}{c|}{\cellcolor{red!5}45.4}          & \cellcolor{red!5}45.3          & \cellcolor{red!5}49.5          & \cellcolor{red!5}42.0          & \cellcolor{green!5}\textbf{43.8} & \cellcolor{green!5}\textbf{27.8}                          & \cellcolor{green!5}\textbf{20.5}                         & \cellcolor{green!5}\textbf{18.5}                                                   \\ \hline
      \end{tabular}}
  \caption{\textbf{One suit of weights} for zero-shot object detection. \textcolor{red!25}{Red} denotes regions where text prompt excels over visual prompt, while \textcolor{green!45}{green} signifies regions favoring visual prompts.}
  \vspace{-1.5em}
  \label{tab:main_result}
  \end{table*}

\noindent\textbf{Data engine for visual prompt.} The training process for visual prompts is to use a portion of the GT box or its center point in the current image as the input. Thus we can leverage established detection datasets including Objects365~\cite{shao2019objects365}, OpenImages~\cite{krasin2017openimages}, HierText~\cite{long2022towards}, CrowdHuman~\cite{shao2018crowdhuman} for the initial training. Meanwhile, to make the data for visual prompt sufficiently diversified, we constructed a data engine to harvest data from SA-1B~\cite{kirillov2023segment}.  This data engine operates through a self-training loop, comprising two phases: \textbf{1) Initial training stage}: In this stage, we first train an initial version of \ourmodel{} with only visual prompt modality on the aforementioned datasets, endowing it with preliminary capabilities for interactive object detection. \textbf{2) Annotation stage}: With the initial model, we then utilize it to annotate the data in SA-1B. SA-1B has tremendous boxes for objects at all granularity. However, the box has no semantic labels, which is not suitable for object detection training. Thus, we employ TAP~\cite{pan2023tokenize} to annotate each box with a category name from a dictionary of 2560 classes. We then adopt the following filtering strategy: if an image has at least one category with a number of instances greater than a certain threshold, it is reserved. However in SA-1B, not all objects have boxes, so we use the original GT box as the interactive visual prompt input and use the initial \ourmodel{} to annotate the missing labeled boxes. In total, we use 2.4M labeled images and 0.65M pseudo-labeled images for visual prompt training.

  \begin{table*}[h]
    \resizebox{\linewidth}{!}{
      \begin{tabular}{ccc|c|cccccccc|cc|c}
        \cline{1-15}
        \multirow{3}{*}{Method} & \multirow{3}{*}{\begin{tabular}[c]{@{}c@{}}Prompt \\ Type\end{tabular}} & \multirow{3}{*}{Backbone} & \begin{tabular}[c]{@{}c@{}}COCO-Val\\ Zero-Shot\end{tabular} & \multicolumn{8}{c|}{\begin{tabular}[c]{@{}c@{}}LVIS \\ Zero-Shot\end{tabular}}                                                                     & \multicolumn{2}{c|}{\begin{tabular}[c]{@{}c@{}}ODinW\\ Zero-Shot\end{tabular}} & \begin{tabular}[c]{@{}c@{}}Roboflow100\\ Zero-Shot\end{tabular} \\ \cline{4-15} 
                                &                                                                         &                           & val-80                                                       & \multicolumn{4}{c|}{minival-804}                                                   & \multicolumn{4}{c|}{val-1203}                                 & \multicolumn{2}{c|}{35val}                                                     & 100val                                                          \\ \cline{4-15} 
                                &                                                                         &                           & AP                                                           & AP            & $AP_f$         & $AP_c$         & \multicolumn{1}{c|}{$AP_{r}$}          & AP            & $AP_f$         & $AP_c$         & $AP_r$         & $AP_{avg}$                                & $AP_{med}$                               & $AP_{avg}$                                                         \\ \cline{1-15}
        T-Rex2                  & Visual-I (Box)                                                          & Swin-T                    & 56.6                                                         & 59.3          & 54.6          & 63.5          & \multicolumn{1}{c|}{64.4}          & 62.6          & 57.3          & 63.7          & 71.9          & 37.7                                   & 39.3                                  & 30.6                                                            \\
        T-Rex2                  & Visual-I (Box)                                                          & Swin-L                    & 58.5                                                         & 62.5          & 57.9          & 66.1          & \multicolumn{1}{c|}{70.1}          & 65.8          & 61.2          & 67.3          & 72.6          & 39.7                                   & 38.1                                  & 30.2                                                            \\
        T-Rex2                  & Visual-I (Point)                                                        & Swin-T                    & 54.3                                                         & 57.4          & 52.1          & 62.3          & \multicolumn{1}{c|}{63.2}          & 60.0          & 54.5          & 60.9          & 68.8          & 34.8                                   & 34.9                                  & 27.7                                                            \\
        T-Rex2                  & Visual-I (Point)                                                        & Swin-L                    & 56.8                                                         & 60.6          & 56.4          & 64.2          & \multicolumn{1}{c|}{65.3}          & 63.8          & 59.1          & 65.1          & 71.1          & 37.5                                   & 35.7                                  & 27.8                                                            \\ \hline
        \end{tabular}}
    \caption{\textbf{One suit of weights} for interactive object detection.}
    \label{tab:main_result_2}
    \end{table*}

\subsection{Model Details}
\ourmodel{} is built upon DINO~\cite{zhang2022dino}. We utilize Swin Transformer~\cite{liu2021swin} as the vision backbone, followed by six layers of transformer encoder layers. We use CLIP-B~\cite{clip} as the text encoder and fine-tune it. For the visual prompt encoder, we stack three layers of deformable cross-attention layer and set the hidden dimension of the feed-forward layer to 1024. We use AdamW~\cite{loshchilov2017decoupled} as the optimizer and set the learning rate to 1e-5 for backbone and text encoder, and 1e-4 for all other modules.

\subsection{Settings and Metrics}
For the object detection task, we evaluate in zero-shot setting, i.e. \ourmodel{} will not be trained on evaluation benchmarks. We report the AP metric on COCO~\cite{lin2014microsoft}, LVIS~\cite{gupta2019lvis}, ODinW~\cite{li2022elevater} and Roboflow100~\cite{ciaglia2022roboflow}. The COCO dataset encompasses 80 common categories. In contrast, the LVIS dataset is characterized by a long-tailed category distribution with 1203 categories. These categories are further segmented into three distinct groups: frequent, common, and rare, with a ratio of \texttt{405:461:337} for the val split, and \texttt{389:345:70} for the minival split~\cite{kamath2021mdetr}. The ODinW and Roboflow100 datasets contain 35 and 100 datasets collected from Roboflow\footnote{https://universe.roboflow.com/}, respectively, covering a variety of scenarios including aerial, video games, underwater, documents, real world, \etc., with long-tailed categories.

We compare several different evaluation protocols for \ourmodel{} under different workflows.

\textbf{Text}: In this protocol, we use all the category names of the benchmark as text prompt inputs, consistent with the previous open-vocabulary object detection setting.

\textbf{Visual-G (Generic)}: In this protocol, \ourmodel{} works on the generic visual prompt workflow. We extract visual prompt embeddings from the training set images of each benchmark for each category. Taking COCO as an example, we first randomly sample  $N$ images for each category that has at least one instance of that category. Next, we extract $N$ visual embeddings for each category using the GT box of each image as input for visual prompting. Subsequently, we compute the average of these $N$ embeddings for each category. These averaged visual embeddings (a total of 80 embeddings) will be used for evaluation. By default, $N$ is set to 16. For each test image, we will repeat this process.

\textbf{Visual-I (Interactive)}: In this protocol, \ourmodel{} works on the interactive visual prompt workflow. Given a test image, suppose it has M categories, then for each category, we randomly select one GT box (or convert it to its center point) in the current image as the visual prompt input for this category. This protocol is relatively easier than Visual-G as we know the category of the test set images in advance, as well as being provided with a GT box. However, despite its simplicity, interactive object detection boasts a wide range of application scenarios, including automatic annotation, object counting, \etc.

\subsection{Zero-Shot Generic Object Detection}
In this study, we explore the zero-shot object detection capabilities of \ourmodel{} across four distinct benchmarks: COCO, LVIS, ODinW, and Roboflow100. The term \textit{zero-shot} refers to the methodological approach where the evaluation benchmarks were not exposed to the model during its training phase, possibly encompassing novel categories and image distributions. As shown in Tab. \ref{tab:main_result}, we observe that text prompt and visual prompt can cover different scenarios respectively. Text prompt demonstrates superior performance in scenarios with relatively common categories. For instance, under the generic visual prompt and Swin-T backbone setting, text prompts surpass visual prompts by a margin of 7 AP points on COCO (80 categories). Similarly, in LVIS-minival (804 categories), text prompts achieve a 5.4 AP point advantage over visual prompts. Conversely, in scenarios characterized by long-tailed distributions, visual prompts exhibit a more robust performance compared to text prompts. Specifically, on LVIS-val (1203 categories), visual prompt leads by 3.4 AP points in the rare group, and by 5.6 AP points on ODinW as well as 9.2 AP points on Roboflow100, underscoring its efficacy in handling less common objects.

\begin{figure}[t]
\centering
\includegraphics[width=1\linewidth]{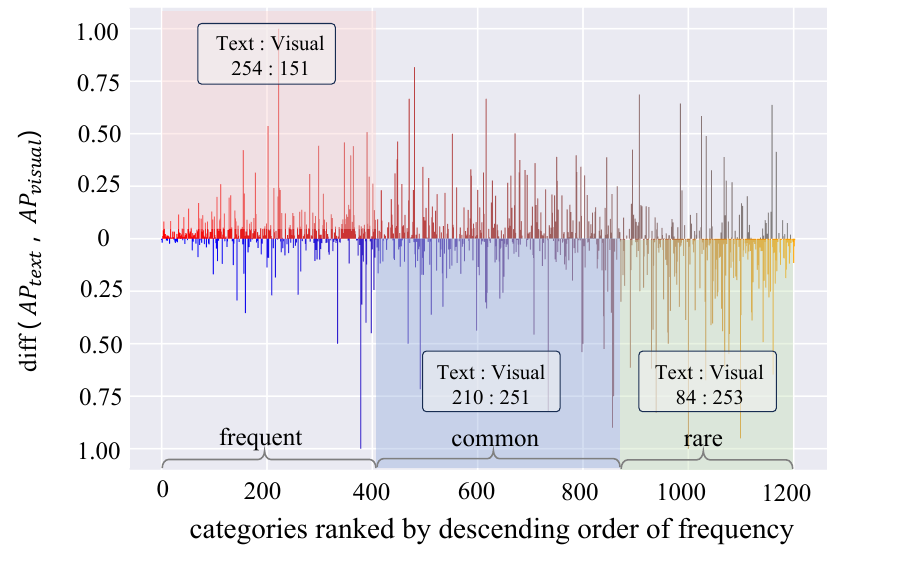}\vspace{-1mm}
\caption{Compare the performance difference between text prompt and visual prompt on the long-tailed dataset LVIS-val. The vertical axis is the difference between the AP of text prompts and visual prompts on each category. The horizontal axis shows the categories of LVIS in descending order of their frequency of occurrence in the text prompt training data. \texttt{Text:Visual} refers to the number of times each has won in the current interval.}
\label{fig:lvis_results} 
\end{figure}

In Fig. \ref{fig:lvis_results}, we show the per-category AP difference between visual prompt and text prompt on the LVIS benchmark. We rank the categories of the LVIS dataset in descending order of their frequency of occurrence in the training set. Our analysis shows that text prompts perform better in recognizing common categories with higher occurrence frequencies. In contrast, visual prompts excel in identifying rarer categories as the frequency decreases. This indicates that text prompts are suited for common concepts, while visual prompts are more effective for rare categories.

\begin{table}[]
  \centering
  \resizebox{0.8\columnwidth}{!}{%
   \begin{tabular}{c|c|c}
\hline
\multirow{2}{*}{Method} & \begin{tabular}[c]{@{}c@{}}FSC147\\ test\end{tabular} & \begin{tabular}[c]{@{}c@{}}FSCD-LVIS\\ test\end{tabular} \\ \cline{2-3} 
                        & MAE                                                   & AP                                                       \\ \hline
FamNet~\cite{ranjan2021learning}                  & 22.08                                                 & -                                                        \\
Counting-DETR~\cite{nguyen2022few}           &                                                       & 22.66                                                    \\
BMNet+~\cite{shi2022represent}                  & 14.62                                                 & -                                                        \\
CountTR~\cite{liu2022countr}                 & 11.95                                                 & -                                                        \\
T-Rex~\cite{jiang2023t}                   & \textbf{8.72}                                         & 40.32                                                    \\ \hline
T-Rex2                  & 10.94                                                 & \textbf{43.35}                                           \\ \hline
\end{tabular}}
\caption{Few-shot object counting results on FSC147~\cite{ranjan2021learning} and FSCD-LVIS~\cite{nguyen2022few} datasets.}
\label{tab:counting}
\end{table}

\subsection{Zero-Shot Interactive Object Detection}
\ourmodel{} also showcases strong interactive object detection capabilities. As shown in Tab. \ref{tab:main_result_2}, the interactive visual prompt significantly outperforms both text prompt and generic visual prompt strategies. However, this comparison may not be entirely equitable, as under the Visual-I setting we have the prior about the categories present in the test image. To provide more insight, we evaluate \ourmodel{} on the few-shot object counting task. In this task~\cite{jiang2023t, nguyen2022few, shi2022represent, liu2022countr}, each test image will be provided with three visual exemplar boxes of the target object and requires to output the number of the target object. We evaluate on FSC147~\cite{ranjan2021learning} and FSCD-LVIS~\cite{nguyen2022few} datasets. Both datasets comprise scenes with densely populated small objects. Specifically, FSC147 typically features single-target scenes, where generally only one type of object is present per image, whereas FSCD-LVIS mainly includes multi-target images. We report the Mean Average Error (MAE) metric for FSC147 and the AP metric for FSCD-LVIS. Following previous work~\cite{jiang2023t}, we use the visual exemplar boxes as the interactive visual prompt. As shown in Tab. \ref{tab:counting}, \ourmodel{} achieves competitive results compared with the previous SOTA algorithm T-Rex. While not matching T-Rex in terms of MAE, \ourmodel{} performs better than T-Rex in terms of AP, which measures the overall detection accuracy. This result suggests that \ourmodel{}’s interactive capabilities are highly capable in dense and small object scenarios.

\subsection{Ablation Experiments}
\noindent\textbf{Ablation of naive joint training.} As demonstrated in Tab. \ref{tab:synergy_ab} (first two rows), the general detection capability of the visual prompt is notably poor (14.0 AP on COCO and 15.3 AP on LVIS-val) when the two prompt modalities are trained separately. The core of the issue lies in the diversity and variance of visual data. For example, when the model is trying to understand what makes a \texttt{chair} when every example the model sees is drastically different from the last. Without a consistent context, it is challenging for the model to form a general concept solely with visual prompts. Upon joint training (second two rows in Tab. \ref{tab:synergy_ab}), the efficacy of visual prompts significantly improves. This improvement suggests that the combination of textual context with visual data helps the model form more stable and generalizable representations. However, the naive joint training without explicit alignment between the two prompts somewhat reduce the effectiveness of text prompts, as both AP on COCO and LVIS dropped.

\begin{figure}[t]
\centering
\begin{subfigure}[b]{0.22\textwidth}
    \centering
    \includegraphics[width=\textwidth]{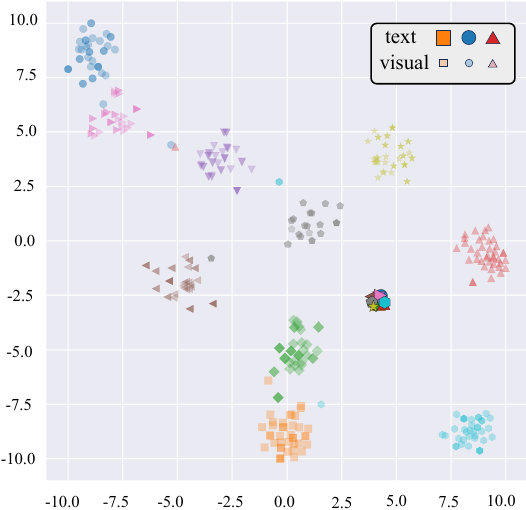}
    \caption{w/o contrastive align}
    \label{fig:tsne_noalign}
\end{subfigure}
\hspace{5pt}
\begin{subfigure}[b]{0.22\textwidth}
    \centering
    \includegraphics[width=\textwidth]{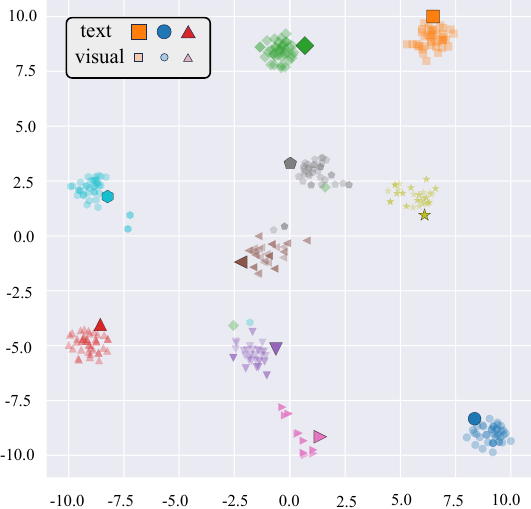}
    \caption{w/ contrastive align}
    \label{fig:tsne_align}
\end{subfigure}
\caption{t-SNE~\cite{van2008visualizing} visualization of text prompt and visual prompt embeddings. We pick the first 10 categories in COCO training set and randomly sample 30 images for each category to get visual prompts.}
\vspace{-1em}
\label{fig:combined}
\end{figure}

The observed decline in text prompt capability could be due to the added complexity of multitask learning. We use t-SNE~\cite{van2008visualizing} to visualize the distribution of text prompt and visual prompt embeddings in Fig. \ref{fig:tsne_noalign}.
We find that the corresponding text prompt and visual prompt embeddings are separated in the feature space, instead of gathered. Therefore the region feature cannot be simultaneously aligned to both the text prompt and the visual prompt, thus making the learning process more challenging.

\begin{table}[]
  \centering
  \resizebox{\columnwidth}{!}{%
    \begin{tabular}{c|c|c|cccc}
  \hline
  \multirow{2}{*}{\begin{tabular}[c]{@{}c@{}}Training\\ Strategy\end{tabular}}         & \multirow{2}{*}{\begin{tabular}[c]{@{}c@{}}Prompt\\ Type\end{tabular}} & \begin{tabular}[c]{@{}c@{}}COCO-Val\\ Zero-Shot\end{tabular} & \multicolumn{4}{c}{\begin{tabular}[c]{@{}c@{}}LVIS-Val\\ Zero-Shot\end{tabular}} \\ \cline{3-7} 
                                                                                       &                                                                        & AP                                                           & AP                                  & $AP_r$                               & $AP_c$                                            & $AP_f$              \\ \hline
  Text Prompt Only                                                                     & Text                                                                   & 46.4                                                         & 32.8                                & 32.1                               & 32.0                                            & 34.0              \\
  Visual Prompt Only                                                                   & Visual-G                                                               & 14.0                                                         & 15.3                                & 8.6                                & 11.3                                            & 22.8              \\ \hline
  \multirow{2}{*}{\begin{tabular}[c]{@{}c@{}}W/O Contrastive\\ Alignment\end{tabular}} & Text                                                                   & 44.4                                                         & 32.2                                & 28.2                               & 28.9                                            & 37.6              \\
                                                                                       & Visual-G                                                               & 38.7                                                         & 30.2                                & 29.4                               & 26.9                                            & 38.7              \\ \hline
  \multirow{2}{*}{\begin{tabular}[c]{@{}c@{}}W/ Contrastive\\ Alignment\end{tabular}}  & Text                                                                   & 45.8\textcolor{deepgreen}{(+1.4)}                            & 34.8\textcolor{deepgreen}{(+2.6)}   & 29.0\textcolor{deepgreen}{(+0.8)}  & 31.5\textcolor{deepgreen}{(+2.6)}               & 41.2\textcolor{deepgreen}{(+3.6)}               \\
                                                                                       & Visual-G                                                               & 38.8\textcolor{deepgreen}{(+0.1)}                            & 34.9\textcolor{deepgreen}{(+4.7)}   & 32.4\textcolor{deepgreen}{(+3.0)}  & 30.3\textcolor{deepgreen}{(+3.4)}               & 41.1\textcolor{deepgreen}{(+2.4)}               \\ \hline
  \end{tabular}}
\caption{Ablationon the proposed text-visual synergy.}
\label{tab:synergy_ab}
\end{table}

\begin{table}[]
  \centering
  \resizebox{\columnwidth}{!}{%
    \begin{tabular}{cc|c|cccc}
\hline
\multirow{2}{*}{\# Prompts} & \multirow{2}{*}{\begin{tabular}[c]{@{}c@{}}Prompt \\ Type\end{tabular}} & \begin{tabular}[c]{@{}c@{}}COCO-Val\\ Zero-Shot\end{tabular} & \multicolumn{4}{c}{\begin{tabular}[c]{@{}c@{}}LVIS-Val\\ Zero-Shot\end{tabular}} \\ \cline{3-7} 
                            &                                                                         & AP                                                           & AP                 & $AP_r$               & $AP_c$               & $AP_f$              \\ \hline
1                           & Visual-G                                                                & 29.2                                                         & 26.2               & 27.6               & 21.3               & 30.9              \\
4                           & Visual-G                                                                & 32.9                                                         & 32.9               & 32.0               & 28.2               & 38.7              \\
16                          & Visual-G                                                                & 38.8                                                         & 34.9               & 32.4               & 30.3               & 41.1              \\
32                          & Visual-G                                                                & 41.3                                                         & 35.1               & 32.2               & 30.3               & 41.7              \\
64                          & Visual-G                                                                & 41.4                                                         & 35.2               & 32.4               & 30.4               & 41.8              \\ \hline
\end{tabular}
  } 
\caption{Ablation experiments on the number of visual prompts and their generic object detection capabilities}
\label{tab:num_visual_promtps}
\end{table}

\begin{table*}[h]
    \resizebox{\linewidth}{!}{
      \begin{tabular}{cc|c|l|c|cccc}
\hline
\multirow{2}{*}{Model} & \multirow{2}{*}{\begin{tabular}[c]{@{}c@{}}Prompt\\ Type\end{tabular}} & \multirow{2}{*}{\begin{tabular}[c]{@{}c@{}}Training\\ Data\end{tabular}} & \multirow{2}{*}{\begin{tabular}[c]{@{}l@{}}Data\\ Size\end{tabular}} & \begin{tabular}[c]{@{}c@{}}COCO-Val\\ Zero-Shot\end{tabular} & \multicolumn{4}{c}{\begin{tabular}[c]{@{}c@{}}LVIS-Minival\\ Zero-Shot\end{tabular}} \\ \cline{5-9} 
                       &                                                                        &                                                                          &                                                                      & AP                                                           & AP             & AP-R            & AP-C          & AP-F                              \\ \hline
Grounding DINO-T       & Text                                                                   & O365, GoldG                                                              & 1.4M                                                                 & 48.1                                                         & 25.6           & 14.14           & 19.6          & 32.2                              \\
Grounding DINO-T       & Text                                                                   & O365, GoldG, Cap4M                                                       & 5.4M                                                                 & 48.4                                                         & 27.4           & 18.1            & 23.3          & 32.7                              \\ \hline
T-Rex2-T               & Text                                                                   & O365, GoldG                                                              & 1.4M                                                                 & 46.1                                                         & 34.9           & 32.7            & 32.9          & 37.1                              \\
T-Rex2-T               & Text                                                                   & O365, GoldG, Bamboo                                                      & 2.5M                                                                 & 45.7                                                         & 38.7           & 35.3            & 39.4          & 38.8                              \\
T-Rex2-T               & Text                                                                   & O365, GoldG, OpenImages, Bamboo, CC3M, LAION                             & 6.5M                                                                 & 46.4                                                         & 39.3           & 35.4            & 40.5          & 39.0                              \\ \hline
T-Rex2-T               & Visual-G                                                               & O365, OpenImages, HierText, CrowdHuman                                   & \multicolumn{1}{c|}{2.4M}                                            & 41.1                                                            & 38.1              & 25.8               & 34.4             & 43.7                                 \\
T-Rex2-T               & Visual-G                                                               & O365, OpenImages, HierText, CrowdHuman, SA-1B                            & \multicolumn{1}{c|}{3.1M}                                            & 38.8                                                         & 37.4           & 29.9            & 33.9          & 41.8                              \\ \hline
T-Rex2-T               & Visual-I (Box)                                                         & O365, OpenImages, HierText, CrowdHuman                                   & \multicolumn{1}{c|}{2.4M}                                            & 41.1                                                         & 40.6           & 40.3            & 43.5          & \multicolumn{1}{l}{38.1}          \\
T-Rex2-T               & Visual-I (Box)                                                         & O365, OpenImages, HierText, CrowdHuman, SA-1B                            & \multicolumn{1}{c|}{3.1M}                                            & 56.6                                                         & 59.3           & 64.4            & 63.5          & \multicolumn{1}{l}{54.6}          \\ \hline
\end{tabular}}
    \caption{Ablation of the proposed data engines.}
    \label{tab:data_engine}
    \end{table*}

\noindent\textbf{Ablations of contrastive alignment.}
As presented in Tab. \ref{tab:synergy_ab} (last two rows), employing contrastive alignment can lead to improved performance for both text and visual prompts. With contrastive alignment, the distribution between text prompt and visual prompt is more structured as shown in Fig. \ref{fig:tsne_align}: text prompts act as anchors and visual prompts cluster around them. This distribution means that visual prompts can learn or derive general knowledge from the closely associated text prompt, making the learning process more efficient. Furthermore, the text prompts are more separated in the feature space compared to Fig. \ref{fig:tsne_noalign}, this indicates that it allows for refinement of text prompts by exposing them to a vast array of visual prompts, thus making them more unique and better defined.

\noindent\textbf{Ablation of generic visual prompt.} In Tab. \ref{tab:num_visual_promtps}, we show that by using more visual prompts, the generic detection capability can be gradually increased. The reason is that visual prompts are not as versatile as text prompts, so we need a large number of visual examples to characterize a generic concept as well as possible. 

\noindent\textbf{Ablation of mixed prompt.}  We further show the results of mixed prompts for generic object detection. This hybrid method aims to leverage the strengths of both modalities to improve detection performance. In Tab. \ref{tab:joint_inference}, the mixed prompt on COCO achieves a result that is balanced between text prompt and visual prompt, while on LVIS there is a further performance improvement. We believe that this hybrid inference workflow is more suitable for the case of long-tailed distributions, where text prompt and visual prompt can promote each other. 

\noindent\textbf{Ablation of data engines.} In Tab. \ref{tab:data_engine}, we ablate the effectiveness of the two data engines. For text prompts, introducing the Bamboo dataset improves the performance on the LVIS dataset (+3.8AP), owing to its diverse categories, but slightly declined performance on the COCO dataset (-0.4AP), indicating that the model is less fitted to COCO categories. Adding image caption data further improves the performance on both benchmarks. For visual prompts, the introduction of the SA-1B data significantly improves the interactive capability of the model, but slightly weakens its generic capability. We speculate that the observed performance degradation may stem from the inadequacy of simply employing TAP~\cite{pan2023tokenize} for object classification within SA-1B, which results in incorrect semantic learning by the model on the SA-1B data. Future work will entail further optimization of this data engine.

\noindent\textbf{Ablation of inference speed.}
In this section, we measure the inference speed of each module of T-Rex2. The experiment is conducted on an NVIDIA RTX 3090 GPU with a batch size of 1. Before measurement, we conducted a warm-up phase to stabilize GPU performance. Inference times were recorded over 100 iterations. The results are shown in Tab. \ref{tab:speed}. Benefiting from the late fusion design, \ourmodel{} can work in real-time when using the interactive visual prompt mode. Specifically, after a user uploads a picture, we only need to process it once with our main processing steps (backbone and encoder) to get the image features. Any further interactions from the user involve just running our visual prompt encoder and decoder multiple times, which is in a real-time manner. This quick response is especially useful for scenarios like automatic annotation.

\begin{table}[]
  \centering
  \resizebox{\columnwidth}{!}{%
    \begin{tabular}{c|ccccc|c|c}
\hline
Backbone & backbone & encoder & \begin{tabular}[c]{@{}c@{}}visual prompt \\ encoder\end{tabular} & \begin{tabular}[c]{@{}c@{}}text prompt \\ encoder\end{tabular} & decoder & FPS   & \begin{tabular}[c]{@{}c@{}}Interactive\\  FPS\end{tabular} \\ \hline
Swin-T   & 0.0318   & 0.0240  & 0.0120                                                           & 0.0103                                                         & 0.0180  & 10.41 & 33.33                                                      \\
Swin-L   & 0.1220   & 0.0929  & 0.0261                                                           & 0.0116                                                         & 0.0240  & 3.62  & 19.96                                                      \\ \hline
\end{tabular}}
\caption{Time cost for each module in \ourmodel{}. Interactive FPS is the inference speed of the visual prompt encoder and the decoder. Since \ourmodel{} is a late fusion model, we only need to forward the backbone and encoder for once, and multi-round interactions only require running the prompt encoder and decoder.}
\label{tab:speed}
\end{table}

\begin{table}[]
\centering
\resizebox{0.8\columnwidth}{!}{%
\begin{tabular}{c|c|cccc}
\hline
\multirow{2}{*}{\begin{tabular}[c]{@{}c@{}}prompt\\ combination\end{tabular}} & \begin{tabular}[c]{@{}c@{}}COCO-Val\\ Zero-Shot\end{tabular} & \multicolumn{4}{c}{\begin{tabular}[c]{@{}c@{}}LVIS-Val\\ Zero-Shot\end{tabular}} \\ \cline{2-6} 
                                                                              & AP                                                           & AP                 & $AP_r$               & $AP_c$               & $AP_f$              \\ \hline
Text                                                                        & \textbf{45.8}                                                & 34.8               & 29.0               & 31.5               & 41.2              \\ \hline
Visual-G                                                                   & 38.8                                                         & 34.9               & 32.4               & 30.3               & 41.1              \\ \hline
\begin{tabular}[c]{@{}c@{}}Text + \\ Visual-G\end{tabular}               & 42.5                                                         & \textbf{37.0}      & \textbf{34.3}      & \textbf{33.8}      & \textbf{41.7}     \\ \hline
\end{tabular}
  }
\caption{Zero-shot object detection results on mixed prompt mode.}
\label{tab:joint_inference}
\end{table}

\section{Conclusion}
\ourmodel{} is a promising attempt towards generic object detection. We reveal the complementary advantages between text prompts and visual prompts, and successfully align the two prompt modalities into a single model, making it both generic and interactive for open-set object detection. We show that these two prompt modalities can benefit from each other and gain performance through contrastive learning. By switching between different prompt modalities in different scenarios, \ourmodel{} demonstrates impressive zero-shot object detection capabilities and can be used in a variety of applications. We hope that this work will bring new insights into the field of open-set object detection and contribute to further development.

\textbf{Limitations. }Despite the integration of text and visual prompts showing mutual benefits within a unified model, challenges arise. Visual prompts may sometimes interfere with text prompts, especially in scenarios involving common objects, as indicated by the reduced performance on the COCO benchmark when both are used together in Tab. \ref{tab:joint_inference}. Despite this, improvements on the LVIS benchmark highlight the potential benefits of this approach. Therefore, further research into improving the alignment between these modalities is essential. Moreover, the requirement for up to 16 visual examples to ensure reliable detection due to visual diversity highlights a need for methods that enable visual prompts to achieve similar effectiveness with fewer visual examples.

\section*{Acknowledgments}
We would like to thank everyone involved in the T-Rex2 project, including project lead Lei Zhang, application lead Wei Liu, product manager Qin Liu and Xiaohui Wang, front-end developers Yuanhao Zhu, Ce Feng, and Jiongrong Fan, back-end developers Weiqiang Hu and Zhiqiang Li, UX designer Xinyi Ruan, and tester Yinuo Chen.
{
    \small
    \bibliographystyle{ieeenat_fullname}
    \bibliography{main}
}

\clearpage

\renewcommand{\thesection}{\Alph{section}} 
\setcounter{section}{0} 
\setcounter{table}{0}
\setcounter{figure}{0}
\maketitlesupplementary

\section{Model Details}
\subsection{Implementation Details}
For the vision backbone, we use Swin Transformer that is pre-trained on ImageNet. For the text encoder, we use the text encoder from the open-sourced CLIP\footnote{\url{https://github.com/openai/CLIP}}. During Hungarian matching, we only use classification loss, box L1 loss, and GIOU loss. The loss weights are 2.0, 5.0, and 2.0, respectively. For the final loss, we use classification loss, box L1 loss, GIOU loss, and constrastive loss, and set the weights to be 1.0, 5.0, 2.0, and 1.0, respectively. Following DINO, we use contrastive denoising training (CDN) to stabilize training and accelerate convergence.

We use automatic mixed precision for training. For the Swin Transformer tiny model, the training is performed on 16 NVIDIA A100 GPUs with a total batch size of 128. For the Swin Transformer large model, the training is performed on 32 NVIDIA A100 GPUs with a total batch size of 64.

\section{Data Engine Details}
\subsection{Text Prompt Data Engine}
To collect region-text pairs from caption datasets LAION400M and Conceptual Captions, We first use CLIP to compute the CLIP score for each image and its caption and retain only pairs of image descriptions with similarity greater than 0.8. Next, we use spaCy to extract the noun phrases in each caption and then use these nouns to prompt the GroundingDINO model to get the box regions corresponding to these noun phrases in the image. Finally, we will compute the CLIP score for each box region and its corresponding noun phrases once more, and keep only the pairs with similarity greater than 0.8.

\subsection{Image Prompt Data Engine}
\begin{figure}[h]
\centering
\includegraphics[width=\linewidth]{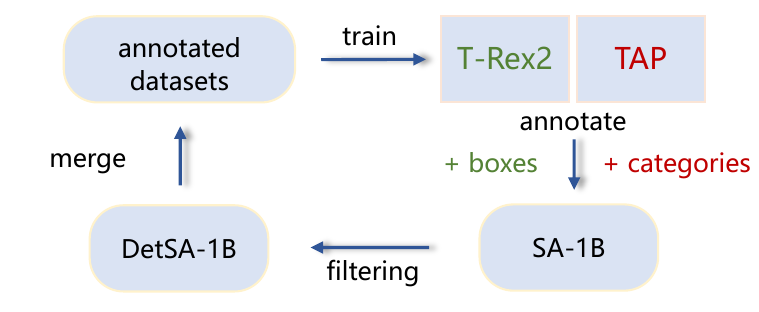}\vspace{-1mm}
\caption{Workflow of the proposed image prompt data engine.}
\label{fig:data_engine}
\vspace{-1mm}
\end{figure}

\begin{figure*}[t]
\centering
\includegraphics[width=\linewidth]{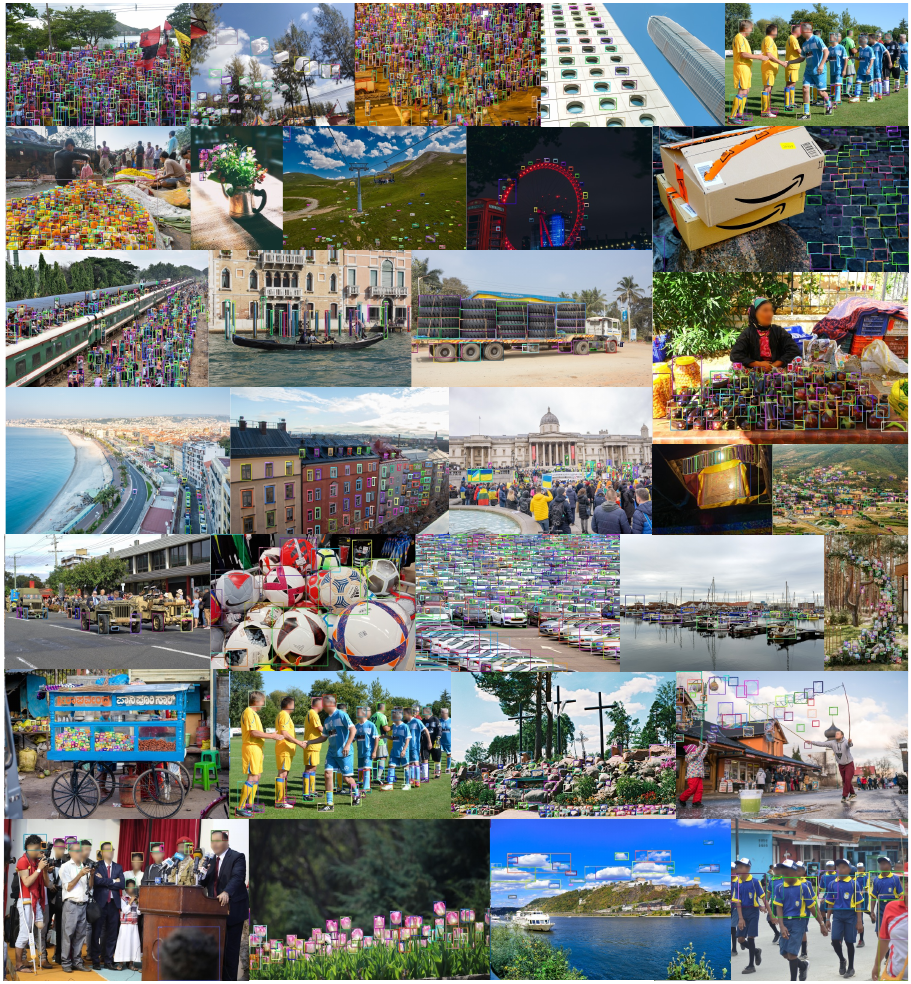}\vspace{-1mm}
\caption{Examples in DetSA-1B.}
\label{fig:detsa_1b}
\vspace{-1mm}
\end{figure*}

We show the overview of the proposed data engine in Fig. \ref{fig:data_engine} and some examples in DetSA-1B in Fig. \ref{fig:detsa_1b}.

\subsection{Data Statistics}

\begin{table}[]
  \centering
  \resizebox{\columnwidth}{!}{%
    \begin{tabular}{c|c|c}
\hline
Data Type                    & Dataset             & \# Images \\ \hline
\multirow{3}{*}{Text prompt} & Conceptual Captions & 1,840,473 \\
                             & LAION400M           & 1,202,245 \\
                             & Bamboo              & 1,109,856 \\ \hline
Visual Prompt                & SA-1B               & 653,285   \\ \hline
\end{tabular}}
\caption{Data statistics of data collected from text prompt and visual prompt engines.}
\label{tab:data_statistics}
\end{table}

We list the amount of data collected from the two data engines in Tab. \ref{tab:data_engine}.
\begin{figure*}[ht]
\centering
\includegraphics[width=\linewidth]{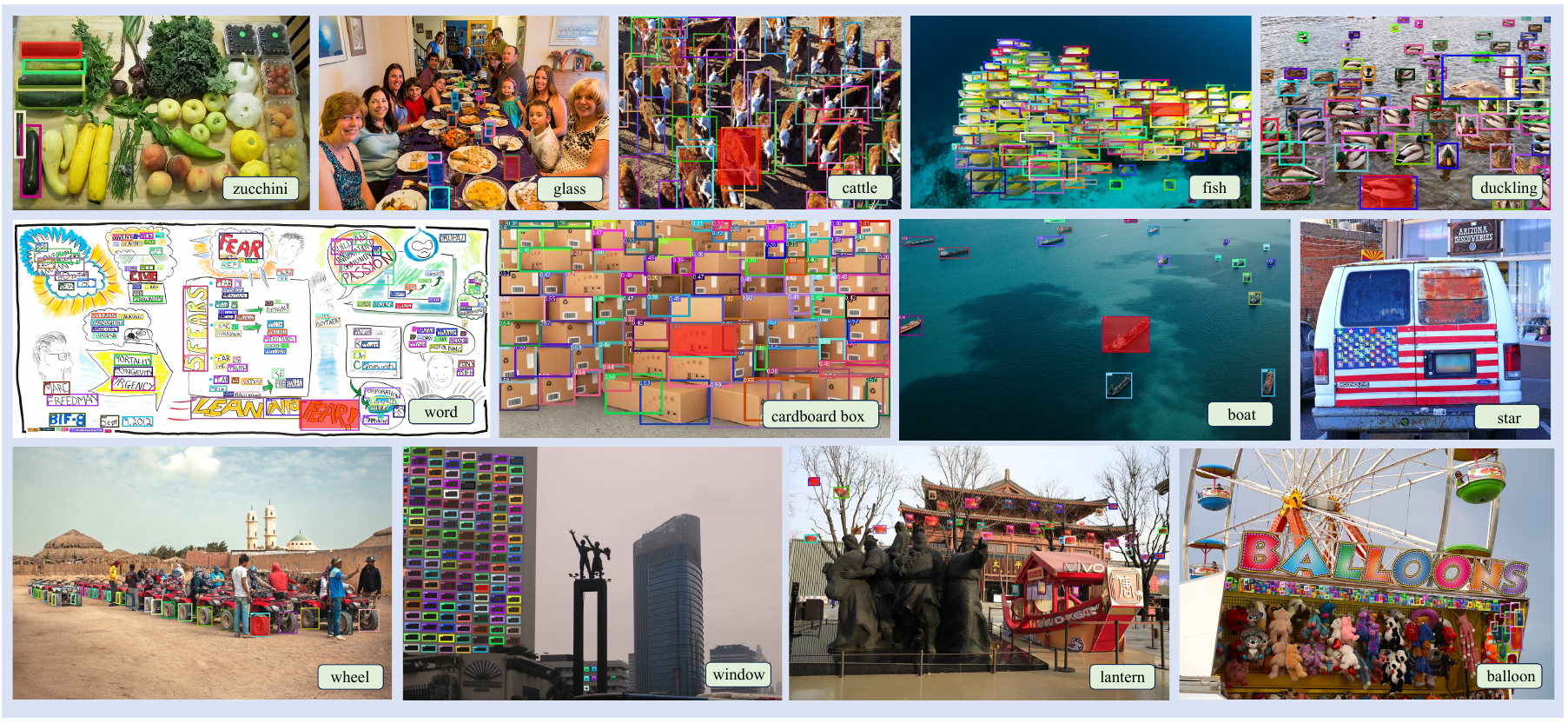}\vspace{-1mm}
\caption{Visualization results of region classification workflow. We use a dictionary of 2560 classes to classify the visual prompts. The classification result is shown at the bottom right for each image.}
\label{fig:region_class}
\vspace{-1mm}
\end{figure*}

\section{Advanced Capabilities for T-Rex2}
\subsection{Region Classification}
\label{sec:Region}
Beyond the aforementioned three inference workflows, \ourmodel{} also supports the capability of region classification. The contrastive alignment between text prompts and visual prompts also unlocks the capability to classify regions for visual prompts. Much like the zero-shot classification approach of CLIP, we can assign category labels to visual prompts by measuring the similarity between visual prompts and pre-computed text prompts:
\begin{equation}
\operatorname{Label} = \operatorname{argmax}j \left(\frac{\exp(V \cdot t_j)}{\sum_{l=1}^K \exp(V \cdot t_l)}\right)
\end{equation}
We can use predefined category names to pre-compute the text embeddings which enable us to identify arbitrary objects through visual prompting. 

We show the zero-shot region classification results on COCO and LVIS in Tab. \ref{tab:region_classification}. We use each GT box as the visual prompt and compute the similarity with all the category names in that dataset. Compared to CLIP, \ourmodel{} possesses stronger region classification capability. We show some visualization results in Fig. \ref{fig:region_class}.

\begin{table}[]
  \centering
  \resizebox{\columnwidth}{!}{%
    \begin{tabular}{cc|cc|cc}
\hline
\multirow{2}{*}{Method} & \multirow{2}{*}{Backbone} & \multicolumn{2}{c|}{\begin{tabular}[c]{@{}c@{}}COCO-Val\\ Zero-Shot\end{tabular}} & \multicolumn{2}{c}{\begin{tabular}[c]{@{}c@{}}LVIS-Val\\ Zero-Shot\end{tabular}} \\ \cline{3-6} 
                        &                           & Acc@Top1                                & Acc@Top5                                & Acc@Top1                                & Acc@Top5                               \\ \hline
CLIP                    & ViT-B                     & 37.6                                    & 60.1                                    & 9.0                                     & 20.0                                   \\ \hline
T-Rex2                   & Swin-T                    & 72.6                                    & 89.4                                    & 40.8                                    & 67.5                                   \\
T-Rex2                   & Swin-L                    & \textbf{82.2}                                    & \textbf{93.9}                                    & \textbf{49.8}                           & \textbf{76.9}                          \\ \hline
\end{tabular}}
\caption{Zero-shot region classification results. For each dataset, we use its full categories as the classification target and calculate the Top1 and Top5 classification accuracy. For CLIP, we crop the region out for classification.}
\label{tab:region_classification}
\end{table}

\subsection{Open-set Video Object Detection}
\begin{figure*}[ht]
\centering
\includegraphics[width=0.8\linewidth]{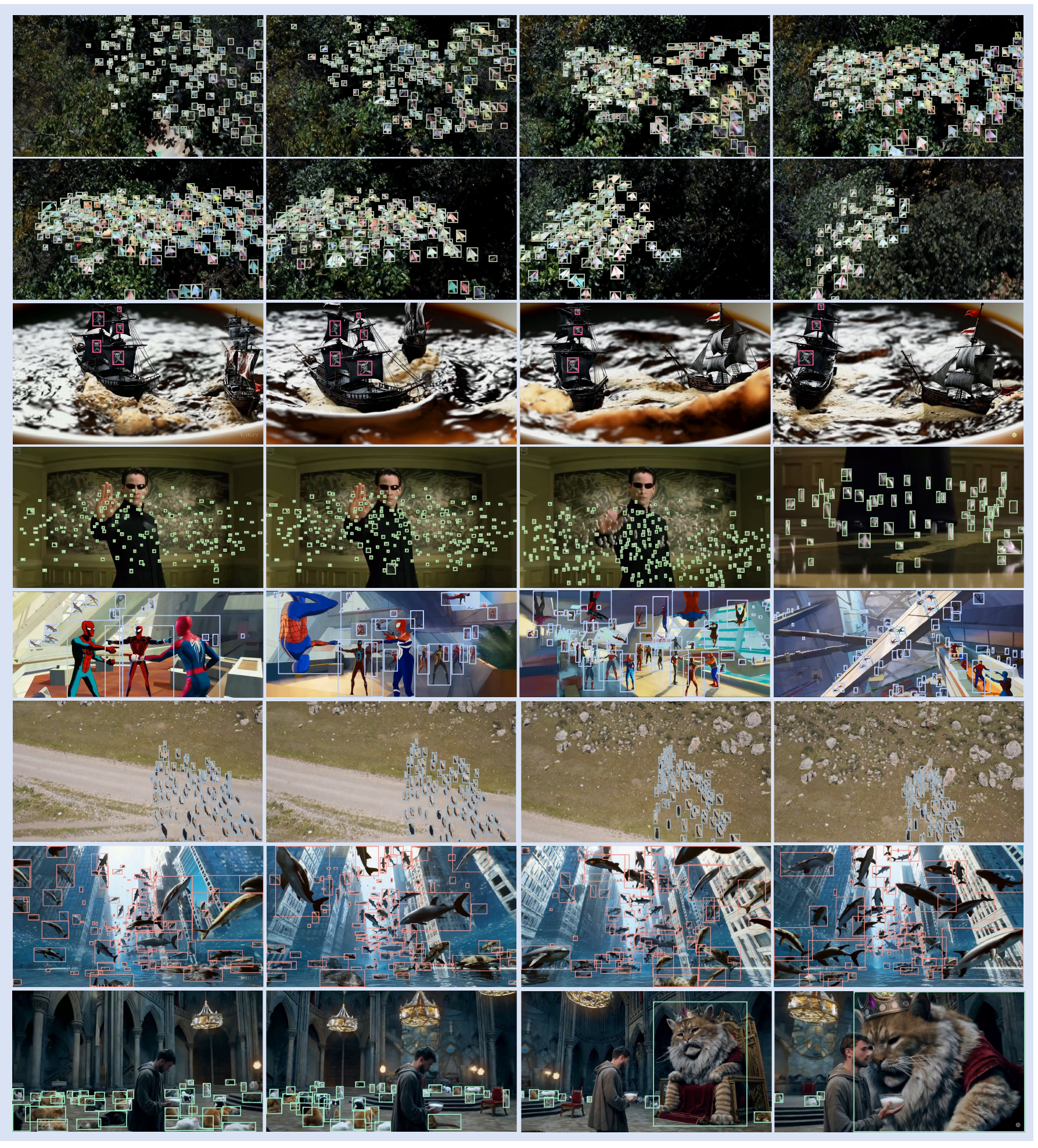}\vspace{-1mm}
\caption{\ourmodel{} on zero-shot video object detection task. We randomly sample 4 frames from a given video and customize a generic visual embedding for an object through the generic visual prompt workflow of \ourmodel{}. This visual embedding will be used for inference for all video frames.}
\label{fig:video_demo}
\vspace{-1mm}
\end{figure*}

\ourmodel{} can also be used for open-set video object detection. Given a video, we can extract any $N$ frames, customize a generic visual embedding for a certain object by using T-Rex2's generic visual prompt workflow, and then use this embedding to detect all frames in the video. We also show some visualization results in Fig. \ref{fig:video_demo}. Despite not being trained on video data, T-Rex2 can also detect objects in videos well. 

\section{More Experiment Details}
\subsection{Details on Object Counting Task}
We evaluate \ourmodel{} on the object counting task to show its interactive object detection capability. Specifically, we are focusing on the few-shot object counting task. In this task, each image will be provided with three exemplar boxes on the current image to indicate the target object and require the output of the number of the target object.

\textbf{Settings.} We conduct evaluations on the commonly used counting dataset FSC147 and the more challenging dataset FSCD-LVIS. FSC147 comprises 147 categories of objects and 1190 images in the test set and FSCD-LVIS comprises 377 categories and 1014 images in the test set. Both
two datasets provide three bounding boxes of exemplar objects for each image, which we will use as the visual prompt for T-Rex2.

\begin{figure*}[t]
\centering
\includegraphics[width=\linewidth]{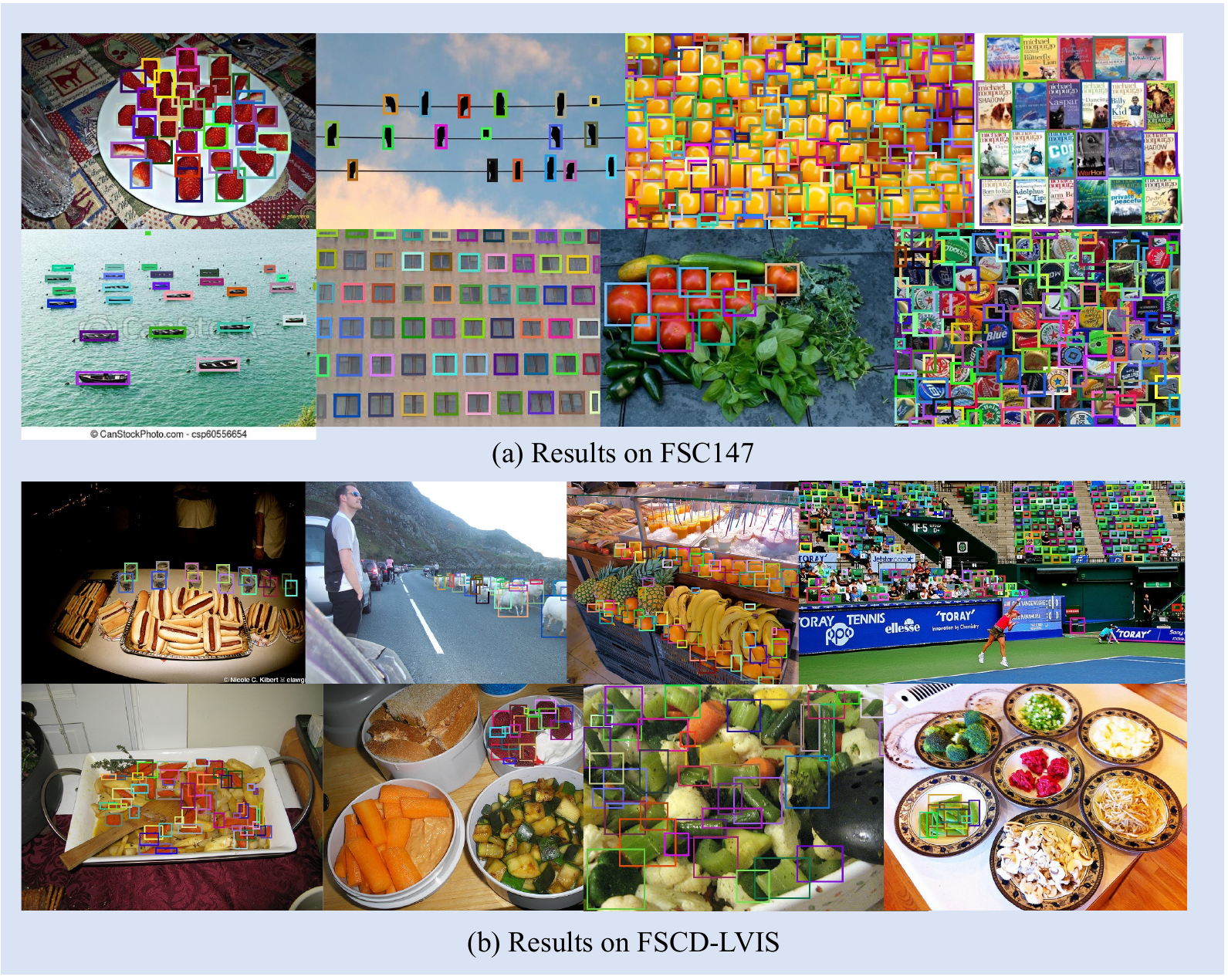}\vspace{-1mm}
\caption{The prediction results of \ourmodel{} on the FSC147 and FSCD-LVIS datasets, respectively.}
\label{fig:counting_vis}
\vspace{-1mm}
\end{figure*}

\textbf{Metric.} We adopt the Mean Average Error (MAE) metric, a widely employed standard in object counting. The mathematical expression is as follows:
\begin{equation}
    \mathrm{MAE}=\frac{1}{J} \sum_{j=1}^J\left|c_j^*-c_j\right|
\end{equation}
We report MAE on the FSC147 dataset as it doesn't provide ground truth boxes on test set images, and report AP on the FSCD-LVIS dataset as it provides ground truth boxes. We show some prediction results of T-Rex2 on the FSC147 and FSCD datasets in Fig. \ref{fig:counting_vis}.

\subsection{Visualization Comparison with T-Rex}
\begin{figure*}[t]
\centering
\includegraphics[width=\linewidth]{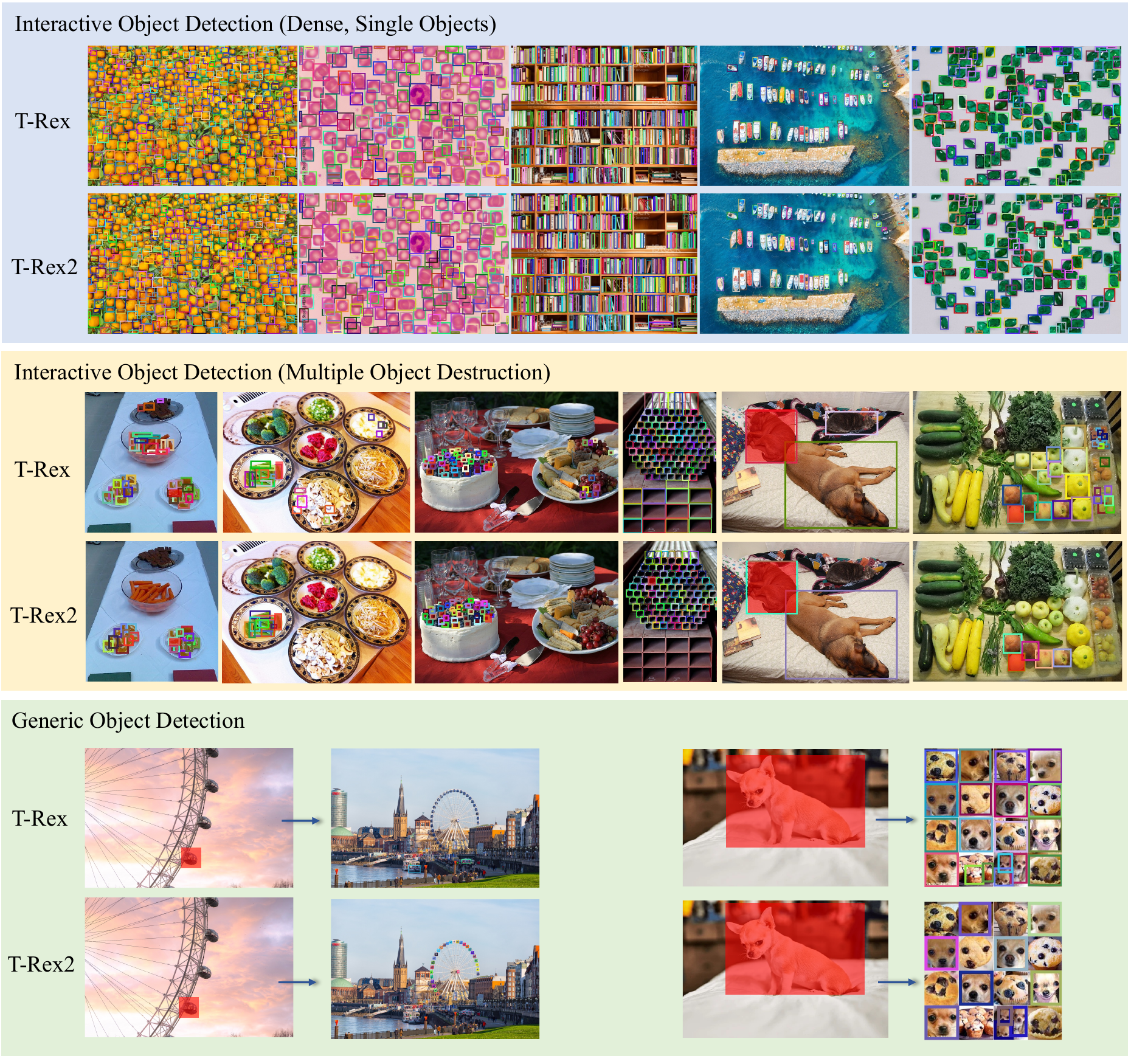}\vspace{-1mm}
\caption{Visualization comparison between T-Rex and \ourmodel{}.}
\label{fig:trex12}
\vspace{-1mm}
\end{figure*}

In Fig. \ref{fig:trex12}, we compare the detection results for T-Rex and \ourmodel{}. In interactive visual prompt detection mode, both models exhibit comparable performance in single-object scenes (where there is no interference from other objects in the image). For multi-object scenarios, T-Rex is more prone to false detections, whereas \ourmodel{} exhibits fewer false detections, indicating a better distinction between objects. This improvement is attributed to the joint training with text and visual prompts. For generic visual prompt detection mode, \ourmodel{} also shows more advantages.

\end{document}